\pdfoutput=1

\documentclass[11pt]{article}

\usepackage{xr-hyper}      
\usepackage{hyperref}
\usepackage{verbatim}

\usepackage{tabularx}
\usepackage{listings}
\usepackage{tcolorbox}
\usepackage{amsmath,amssymb}

\usepackage[final]{acl}

\usepackage{times}
\usepackage{latexsym}
\usepackage{array} 
\usepackage{amsfonts} 

\usepackage[T1]{fontenc}

\usepackage[utf8]{inputenc}

\usepackage{microtype}

\usepackage{inconsolata}

\usepackage{graphicx}

\title{Seeing Through Words, Speaking Through Pixels: Deep Representational Alignment Between Vision and Language Models}

\author{
  Zoe Wanying He\textsuperscript{} \quad
  Sean Trott\textsuperscript{} \quad
  Meenakshi Khosla\textsuperscript{} \\
  Department of Cognitive Science \\
  University of California, San Diego \\
  \texttt{\{wah016, sttrott, mkhosla\}@ucsd.edu}
}

\begin{document}
\maketitle
\begin{abstract}
Recent studies show that deep vision-only and language-only models—trained on disjoint modalities—nonetheless project their inputs into a partially aligned representational space. Yet we still lack a clear picture of \emph{where} in each network this convergence emerges, \emph{what} visual or linguistic cues support it, \emph{whether} it captures human preferences in many-to-many image-text scenarios, and \emph{how} aggregating exemplars of the same concept affects alignment. Here, we systematically investigate these questions. We find that alignment peaks in mid-to-late layers of both model types, reflecting a shift from modality-specific to conceptually shared representations. This alignment is robust to appearance-only changes but collapses when semantics are altered (e.g., object removal or word-order scrambling), highlighting that the shared code is truly semantic. Moving beyond the one-to-one image-caption paradigm, a forced-choice "Pick-a-Pic" task shows that human preferences for image-caption matches are mirrored in the embedding spaces across all vision-language model pairs. This pattern holds bidirectionally when multiple captions correspond to a single image, demonstrating that models capture fine-grained semantic distinctions akin to human judgments. Surprisingly, averaging embeddings across exemplars amplifies alignment rather than blurring detail. Together, our results demonstrate that unimodal networks converge on a shared semantic code that aligns with human judgments and strengthens with exemplar aggregation.
\end{abstract}

\section{Introduction}

\begin{figure*}[!t]
\begin{center}
\includegraphics[width=0.9\textwidth]{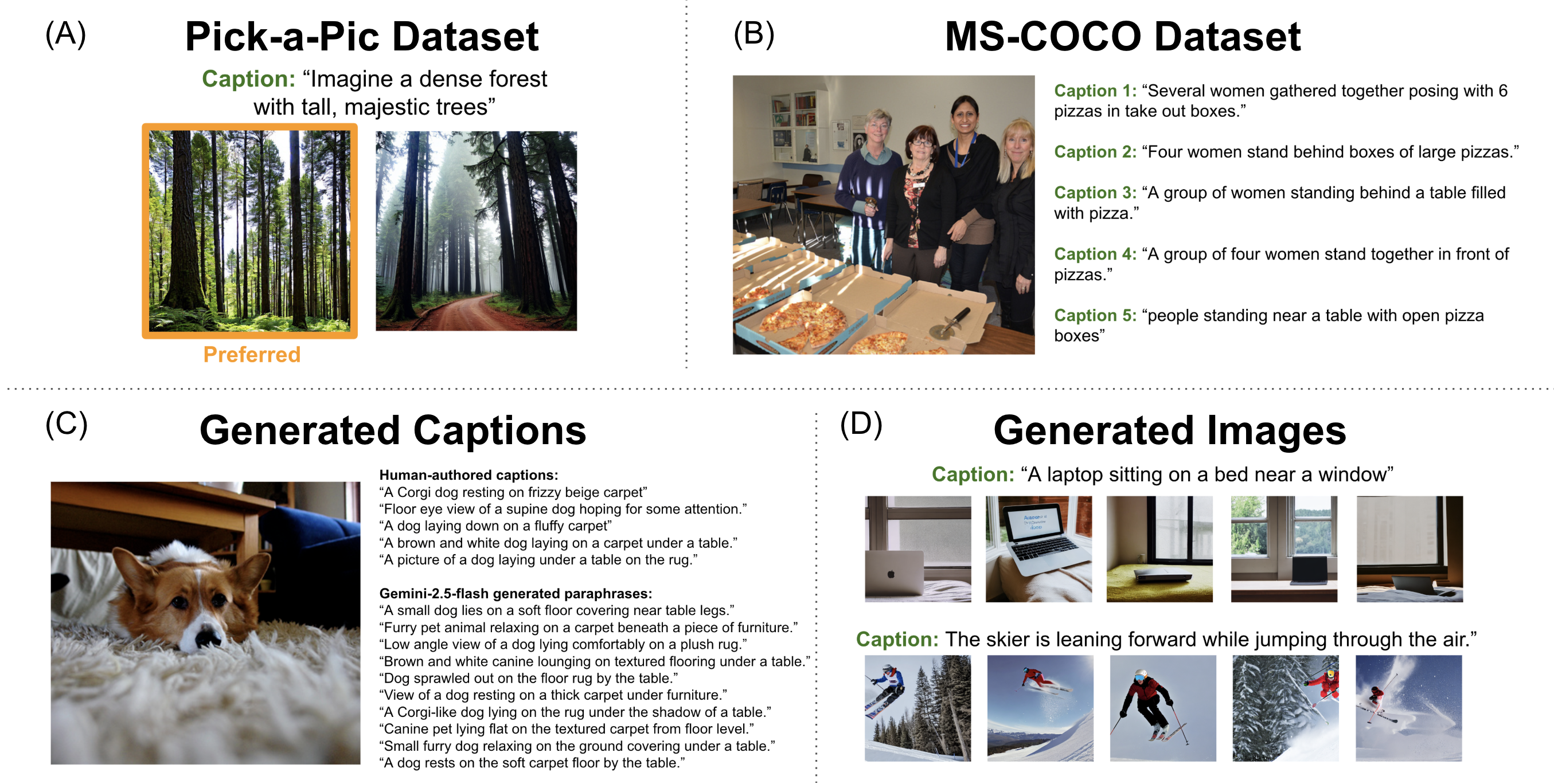}
\caption{Example data from \textbf{(A)} the Pick-A-Pic dataset and \textbf{(B)} the MS-COCO dataset. \textbf{(C)} Example captions generated by Gemini-2.5-Flash by paraphrasing the human-authored captions in MS-COCO. \textbf{(D)} Example MS-COCO captions and synthesized images by the stable-diffusion model.}
\label{fig:data}
\end{center}
\end{figure*}


The idea of a universal, modality‐independent substrate of meaning has long intrigued philosophy, neuroscience and cognitive science—from Plato’s ideal forms to Fodor’s “Language of Thought” (mentalese) \cite{fodor1975language}. This motivates a foundamental question: do putatively distinct systems—such as vision and language models—encode meaning in a shared, abstract space or in modality‐specific codes?

Rapid developments in AI—particularly large-scale vision and language models—provide novel tools to explore these ideas computationally. Large-scale vision-only and language-only models, trained on massive but disjoint corpora, nonetheless exhibit striking representational convergence. \citet{huh2024platonic} coined this phenomenon the "Platonic Representation Hypothesis", showing that increasingly capable LLMs align more tightly with larger vision models. Interestingly, this alignment occurs even without explicit cross-modal training. This "Platonic Representation Hypothesis" is further supported by \citet{maniparambil2024vision}, who demonstrate that this convergence manifests across a range of model architectures and training paradigms~. 

Critically, cross-modal alignment is not merely correlational. \citet{merullo2022linearly} show that training just one linear projection is enough to map a frozen vision-transformer’s embeddings into the token-embedding space of a frozen language model, letting the stitched system caption images and answer visual questions without any additional multimodal training. Similarly, \citet{koh2023grounding} show analogous gains for the reverse mapping from text to image, showing that a frozen LLM can be visually grounded with a single learned linear map, achieving strong zero-shot performance on tasks such as contextual image retrieval and multimodal dialogue. 

\citet{marjieh2024large} show that even multimodal models like GPT-4 rely predominantly on textual associations rather than direct visual input when predicting human perceptual judgments—highlighting language as a sufficient scaffold for grounding sensory semantics. \citet{bavaresco2025experiential} demonstrate that text alone—when modeled on scale—can implicitly encode rich experiential semantics, echoing \citet{marjieh2024large}'s results on LLMs’ ability to recover perceptual hierarchies like the pitch spiral.

Convergent evidence also emerges from neuroscience. \citet{popham2021visual} used within-subject fMRI to chart voxel-wise semantic tuning during silent-movie viewing (purely visual) and narrative listening (purely linguistic). They discovered that the two modality-specific maps are topographically contiguous: for every visual category encoded in posterior occipital cortex, a mirror linguistic representation appears immediately anterior to the same cortical border. In other words, visual and linguistic semantics form a single, smoothly joined map that straddles the edge of human visual cortex, implying a tightly aligned cross-modal code rather than two isolated systems. \citet{doerig2022semantic} asked whether vision already encodes such linguistic semantics. They showed that a vision model trained to translate images directly into sentence embeddings of a language model predicts voxel patterns even better than the embeddings themselves, offering a mechanistic account of how the visual system may recast images into a language-like semantic code by default. \citet{saha2024modeling} went further, finding that off-the-shelf LLM embeddings sometimes outperform dedicated vision models in explaining activity in high-level visual areas. Together, these findings suggest that the cross-modal alignment observed in artificial networks may reflect, or even recapitulate, the brain’s own amodal semantic code.

These findings collectively suggest that modern vision and language models, and possibly even brain systems—like Plato’s ideal forms—incrementally discard modality-specific details in favor of a shared, amodal semantic code.

Yet critical gaps remain. First, where along the network hierarchy does this alignment emerge, and is it symmetric across modalities? Second, what visual attributes or linguistic properties drive the effect? Third, all previous demonstrations of cross-modal alignment rely on one-to-one image–text pairs. These analyses inadvertently mask the complexity of real-world semantics where no single description exhausts an image’s meaning, and the same sentence can fit many images.

In this study, we fill these gaps through extensive analyses of cross-modal alignment on a broad suite of vision and language models. We map alignment layer-by-layer and probe its dependence on targeted manipulations—semantic (object removal, role shuffling) versus appearance-only. Alignment peaks in mid-to-late layers of both modalities, collapses under semantic changes, and is largely unaffected by superficial appearance edits.

To address the third gap about the many-to-many mapping between images and text, our study employs two complementary analyses that explicitly investigate semantic alignment at a finer granularity using many-to-many mappings. First, using a forced-choice "Pick-a-Pic" task, we show that visual embeddings of human-preferred images align more closely with the language model embeddings of the caption than non-preferred images. Second, for the same image, we analyze pairs of captions selected based on high and low CLIP-scores—previously validated as proxies for human preferences—and observe analogous alignment patterns. These results indicate that vision and language models converge on a common semantic ground that reflects subtle distinctions aligned with human judgments. 

In our second analysis, we investigate the impact of aggregating embeddings across multiple images associated with a single caption and vice versa. Contrary to the intuitive expectation that averaging embeddings would diminish representational specificity, we discover that such aggregation consistently enhances alignment. This suggests that rather than blurring distinctions, averaging distills a more stable, modality-independent semantic core shared across representations. Together, our findings reveal that examining many-to-many correspondences offers richer insights into cross-modal alignment, highlighting a robust convergence toward a shared conceptual space that captures subtle and complex semantic relationships.

We summarize our contributions as follows:
\begin{itemize}
  \item \textbf{Layer-wise alignment.} Alignment strengthens with depth and converges toward a shared space, asymmetrically across modalities.
  \item \textbf{Semantic sensitivity.} Alignment drops under semantic edits but is robust to appearance-only changes.
  \item \textbf{Human consistency.} The aligned space reflects human preferences (Pick-a-Pic; CLIP-ranked captions).
  \item \textbf{Embedding aggregation.} Averaging embeddings across captions/images improves alignment, indicating a modality-independent semantic core.
\end{itemize}


    
    

\begin{figure*}[t]
\begin{center}
\includegraphics[width=0.9\textwidth]{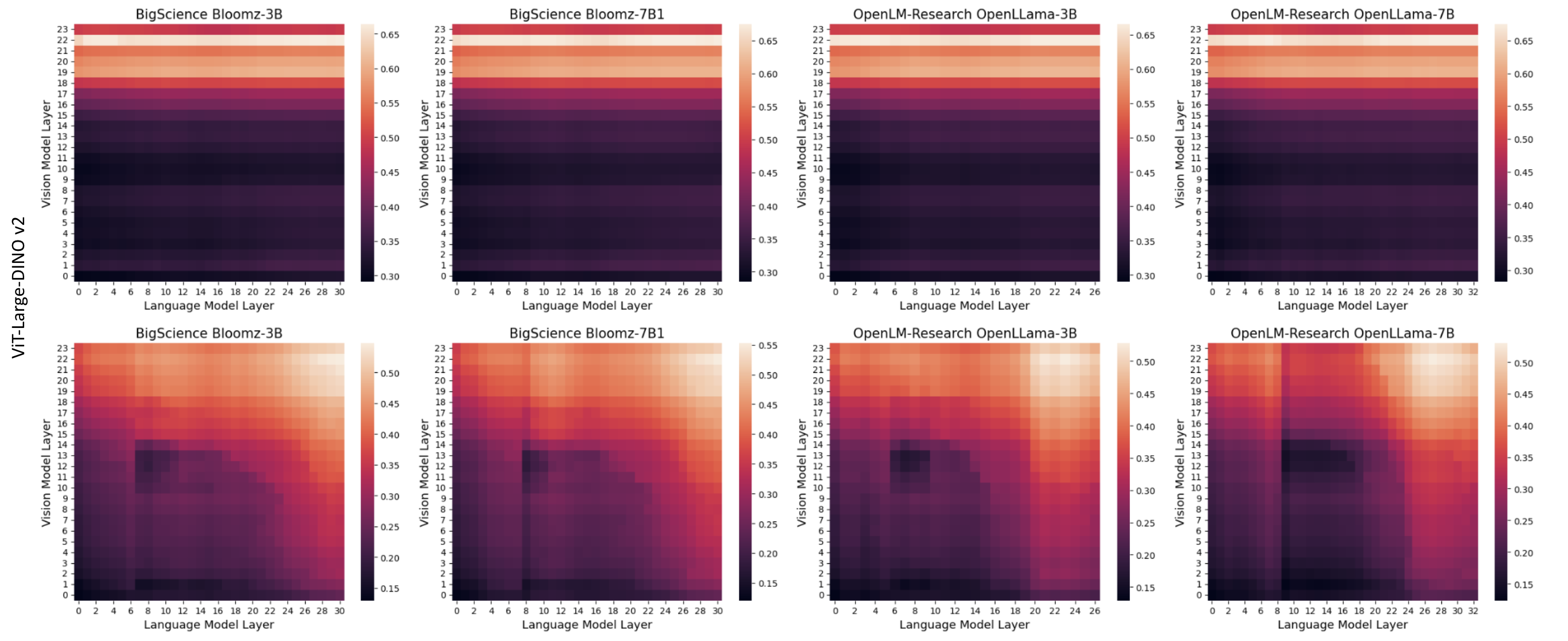}
\caption{Layer-wise alignment, measured suing linear predictivity score, between one example vision model (ViT-Large-Dinov2) and all language models. Top row: Alignment computed in language-to-vision direction. Bottom row: Alignment computed in vision-to-language direction.}
\label{fig:heatmap}
\end{center}
\end{figure*}

\section{Methods}
We compare image representations from large vision models with textual representations of the same images from large language models. For vision models, we employed Vision Transformers (ViTs) trained via DINOv2 \cite{oquab2023dinov2} on the LVD-142M dataset.  DINOv2 learns rich visual representations by solving a self-distillation task where a student network is trained to match the output distribution of a teacher network (an exponential moving average of the student) while viewing different augmented versions of the same image. For language models, we employed BLOOM~\cite{bigscience2022bloom}, a decoder-only transformer-based architecture trained on a massive multilingual corpus, and OpenLLaMA, an open-source reproduction of the LLaMA model trained on publicly available datasets\cite{GengLiu2023}. Multiple model sizes were selected from repositories including Huggingface \cite{wolf2019huggingface} and PyTorch Image Models (TIMM) \cite{Wightman2021}.

Beyond the primary models, we evaluate three additional LLM families (Qwen, Phi-3, SmolLM) for all core analyses; see Appendix~\ref{app:expanded-llm}.

For images, the class token from the penultimate transformer block is used; for language, token activations are averaged from the same layer.


Two datasets are employed:
\begin{itemize}
    \item \textbf{Pick-A-Pic:} An open dataset of over 500K human preference judgments on text-to-image outputs, collected from 37K real-user prompts; each prompt has two generated images and a binary/tie preference label \cite{Kirstain2023}. Here, we randomly sample 1,000 prompt–image-pair judgments for analysis (Figure ~\ref{fig:data}A).
     \item \textbf{MS-COCO:} An image captioning dataset of 123K natural images depicting complex everyday scenes, each annotated with five human-authored captions \cite{lin2014microsoft}. Here, we randomly sample 1,000 images (and their associated captions) from the official validation split (Figure ~\ref{fig:data}B). 
\end{itemize}


To assess dataset generalization, we also replicate on Flickr8k for the core analyses; see Appendix~\ref{app:addl-datasets-brief}.

\subsection*{Computing Alignment}
To quantify alignment between representations from language and vision models, we use \textit{linear predictivity}. For each pair of representations, $\mathbf{X} \in \mathbb{R}^{n \times d_X}$ (e.g., from a vision model) and $\mathbf{Y} \in \mathbb{R}^{n \times d_Y}$ (e.g., from a language model), we fit a ridge regression from $\mathbf{X}$ to $\mathbf{Y}$:

\begin{equation}
\hat{\mathbf{W}} = \arg\min_{\mathbf{W}} \|\mathbf{X}\mathbf{W} - \mathbf{Y}\|^2_2 + \lambda \|\mathbf{W}\|^2_F,
\end{equation}

with $\lambda$ chosen by cross-validation over $10^{-8}\!\dots\!10^{8}$.  Alignment is the average Pearson correlation between predicted and actual responses across all units and five cross-validation folds. 


We treat this as an \textit{asymmetric} similarity measure and report results for both directions: predicting language representations from vision ($\mathbf{X} \rightarrow \mathbf{Y}$) and vice versa ($\mathbf{Y} \rightarrow \mathbf{X}$). This allows us to disentangle directional differences in information content across modalities.

For metric robustness, core analyses are repeated with \emph{CKA} (Appendix~\ref{app:metrics}) on the same held-out items and layers (Appendix~\ref{app:cka}).




\section{Results}

We examine (i) layer-wise alignment, (ii) input manipulations, (iii) human preference, and (iv) embedding averaging. Replications hold under CKA and on Flickr8k within scope, and across additional LLM families; see Appendix~\ref{app:cka},~\ref{app:addl-datasets-brief},~\ref{app:expanded-llm}.

\subsection{Layer-wise vision-language alignment}


To pinpoint where vision–language alignment first appears and how it evolves across the network hierarchy, we performed a layer-by-layer mapping between each pair of vision-transformer and language-model embeddings. As shown in Figure \ref{fig:heatmap}, both modalities exhibit low cross-modal predictivity in their earliest layers and increase through the mid and later layers. These patterns hold consistently across different vision-language model pairs (Figure \ref{fig:heatmap_new}). These findings demonstrate that both vision and language models transition from modality-bound encoding toward an abstract, shared semantic space as depth increases. 

We also observe a clear directional asymmetry in these mappings. When mapping from language to vision, we find that even early language layers can successfully predict later vision layers. In contrast, mapping from vision to language reveals a more graded effect: deeper vision layers progressively yield higher predictivity for deeper language layers. Early vision features poorly predict any language layer, while later vision representations align best with higher language layers. This asymmetry suggests that textual representations abstract away from surface form more rapidly than visual ones, while vision networks require deeper processing to reach a comparable semantic level. 


\subsection{Semantic content, not surface form, drives cross-modal alignment}
\begin{figure*}[t]
\begin{center}
\includegraphics[width=0.96\textwidth]{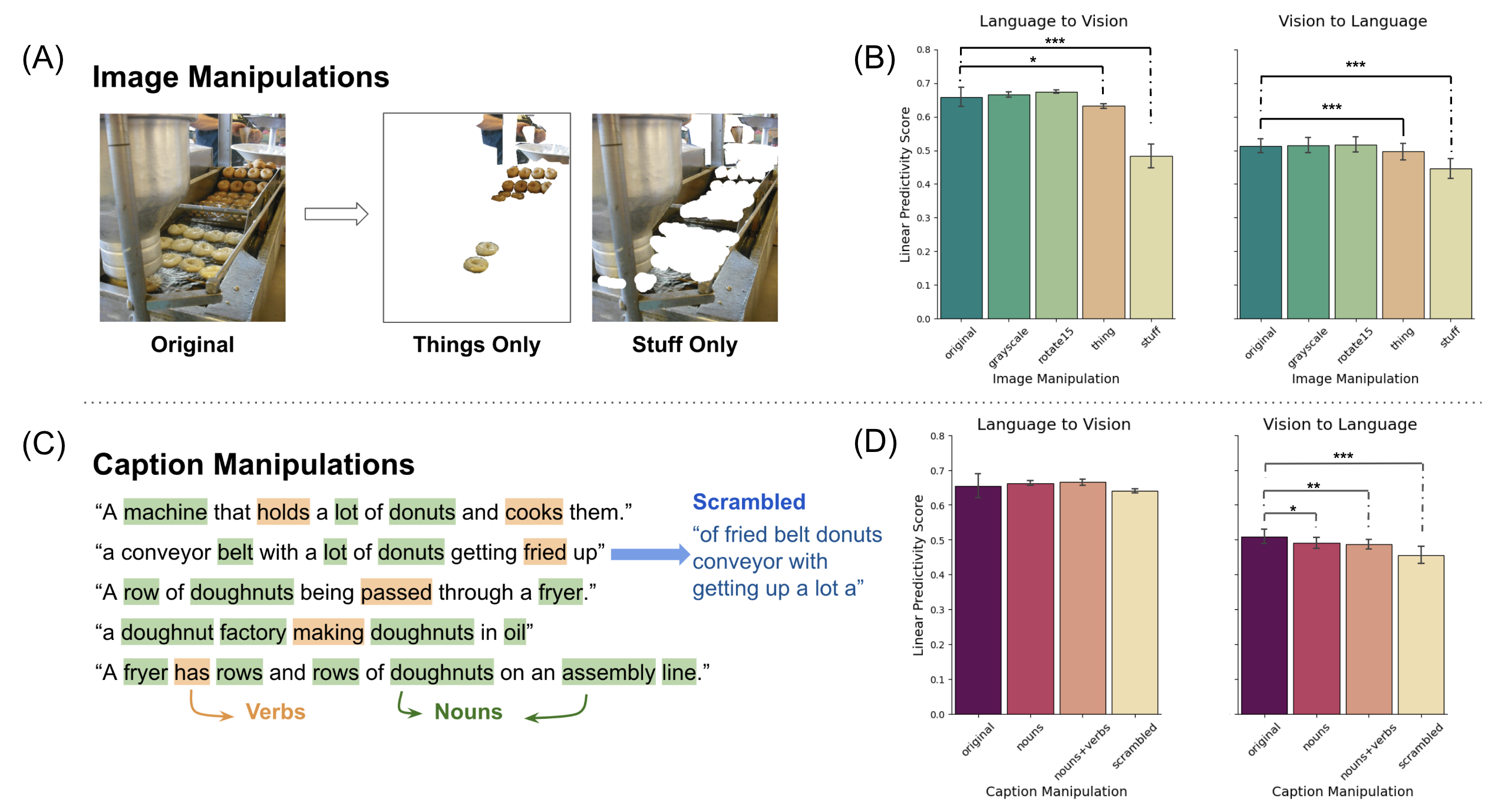}
\caption{\textbf{(A)} Example thing-only and stuff-only images by manipulating the original image using masks from COCO-Stuff. \textbf{(B)} Alignment by image manipulations. \textbf{(C)} Demonstration of image manipulations: nouns and verbs extraction, and captions scrambling. \textbf{(D)} Alignment by caption manipulations. Paired t-tests (n=8 vision-language model pairs per comparison) were conducted separately for each image manipulation, and p-values were adjusted for four comparisons per mapping direction using the Benjamini–Hochberg procedure (FDR).}
\label{fig:input_manipulations}
\end{center}
\end{figure*}



We next explore whether the cross-modal correspondence we observe is mainly driven by surface form or by deeper semantic content. 

\subsubsection{Image manipulations} To dissociate appearance-level similarity from semantic correspondence, we performed four controlled perturbations on each MS-COCO image. Two manipulations altered only the appearance while preserving the full meaning: (i) conversion to grayscale and (ii) 15 degree image rotation. The other two manipulations altered the semantic content with different degrees by exploiting the segmentation masks (Figure~\ref{fig:input_manipulations}A) provided with COCO-Stuff ~\citep{caesar2018coco}:
\begin{itemize}
    \item \textbf{Thing-only views} that preserve instances of the foreground object classes (e.g. person, car) but remove the surrounding context to eliminate spatial and contextual relations;
    \item \textbf{Stuff-only views} that retain only the background layout and the scene categories (e.g. grass, wall) while removing the foreground objects. 
\end{itemize}
We find that appearance-only manipulations of image inputs have no notable negative effects on alignment (Figure~\ref{fig:input_manipulations}B, grayscale: L→V: $t(7)=-0.8405$, $p=0.4284$, $q=0.4284$; V→L: $t(7)=-1.3386$, $p=0.2226$, $q=0.2543$; rotation: L→V: $t(7)=-1.7569$, $p=0.1224$, $q=0.1631$; V→L: $t(7)=-3.1161$, $p=0.0169$, $q=0.0271$). In contrast, deleting semantic content from images results in substantial alignment degradation (Figure ~\ref{fig:input_manipulations}B). Isolating only the foreground “thing” pixels and removing contextual relations significantly lowered the alignment scores (L→V: $t(7)=3.4304$, $p=0.0110$, $q=0.0220$; V→L: $t(7)=7.2528$, $p=0.0002$, $q=0.0005$). Retaining only the “stuff” background further reduced the alignment (L→V: $t(7)=10.1267$, $p<0.0001$, $q=0.0001$; V→L: $t(7)=11.7109$, $p<0.0001$, $q=0.0001$).

Notably, the decline was systematically steeper in the language-to-vision direction, indicating that mapping from textual embeddings to visual layers depends more heavily on intact visual semantics.

\subsubsection{Caption manipulations}
To explore the linguistic properties driving the alignment, we separately manipulated the captions in the MS-COCO dataset with different levels of semantic disruption by retaining: (i) nouns only, (ii) nouns and verbs, and (iv) all the words but in scrambled order (Figure~\ref{fig:input_manipulations}C; Appendix~\ref{app:caption_manipulation}). 

Interestingly, only in the vision-to-language mapping direction do caption manipulations negatively affect the alignment (Figure~\ref{fig:input_manipulations}D, right). Specifically, nouns-only ($t(7)=3.5956$, $p=0.0088$, $q=0.0176$) and nouns+verbs ($t(7)=5.3561$, $p=0.0011$, $q=0.0032$) show similar moderate decreases, while scrambled captions produce the largest drop ($t(7)=22.8176$, $p<0.0001$, $q<0.0001$). This suggests that nouns and verbs carry the primary semantic weight in grounding language to visual content, while word order and the full lexical distribution become even more crucial when projecting from vision to language.

The directional asymmetries we observe—greater sensitivity of language→vision mapping to intact visual semantics and of vision→language mapping to linguistic composition—suggest complementary organizational principles in how each modality abstracts and transmits meaning across the shared representational space. 








\subsection{Vision–language alignment mirrors human preferences}

\begin{figure}[ht]
\begin{center}
\includegraphics[width=0.48\textwidth]{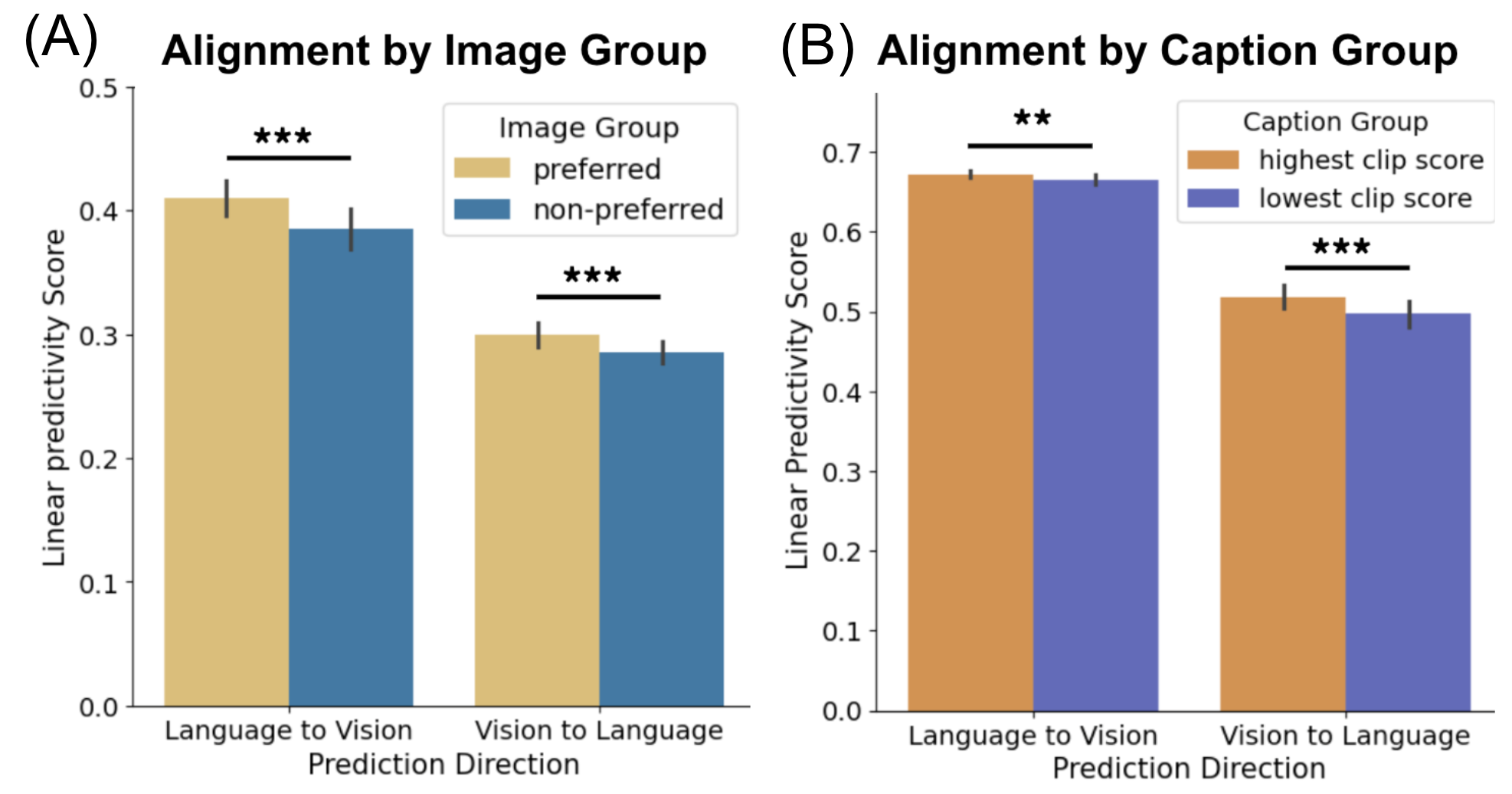}
\end{center}
\caption{\textbf{(A)} Pick-a-Pic dataset Linear predictivity scores grouped by image variation (preferred vs. non-preferred) based on human judgments. \textbf{(B)} MS-COCO dataset Linear predictivity scores grouped by caption variation based on CLIP Scores. Error bars indicating the standard error across model pairs.} 
\label{fig:bar}
\end{figure}

We next evaluated whether cross-modal alignment tracks fine-grained human preferences. Images from the ``Pick-a-Pic'' dataset, which provides two generated images for the same prompt with human preference judgments, were grouped into high- and low-preference categories. For each group, vision model representations were extracted and linear predictivity scores were computed using the corresponding caption embeddings. This design probes alignment at a finer-grained resolution: can the vision–language mapping replicate the subtle distinctions that lead people to prefer one image over another, even when the linguistic description is identical?

Our results indicate that images preferred by human raters exhibit significantly stronger alignment with their associated captions than non-preferred images across all vision-language model pairs (paired t-test, L→V: $t(7)=19.8225$, $p<0.001$; V→L: $t(7)=10.2338$, $p<0.001$; Figure ~\ref{fig:bar}A).  In other words, even when two pictures illustrate the same text, the uni-modal vision and language models collectively ``agree'' with human raters about which picture is the better semantic fit. This fine-grained sensitivity shows that the cross-modal alignment we measure is not a coarse correlation but captures subtle, human-relevant distinctions within a shared semantic space.

A complementary analysis from the text side reinforces this conclusion. We computed the CLIP Score \cite{hessel2021clipscore}—a reference-free metric based on the cosine similarity of image–caption embeddings—for all MS-COCO captions, as a reasonable proxy for human preferences \cite{hessel2021clipscore}. Our analysis reveals that captions with higher CLIP scores are significantly more aligned with their images than those with lower scores (paired t-test, language-to-vision: $t(7)=3.9231$, $p=0.0057$; vision-to-language: $t(7)=17.8350$, $p<0.001$; Figure ~\ref{fig:bar}B).

Thus, across seven vision–language model pairs evaluated on MS-COCO and Pick-a-Pic, the model embeddings capture fine-grained semantic distinctions that mirror human evaluative patterns.

\subsection{Averaging embeddings across multiple captions and images enhances alignment}

\begin{figure}[ht]
\begin{center}
\includegraphics[width=0.48\textwidth]{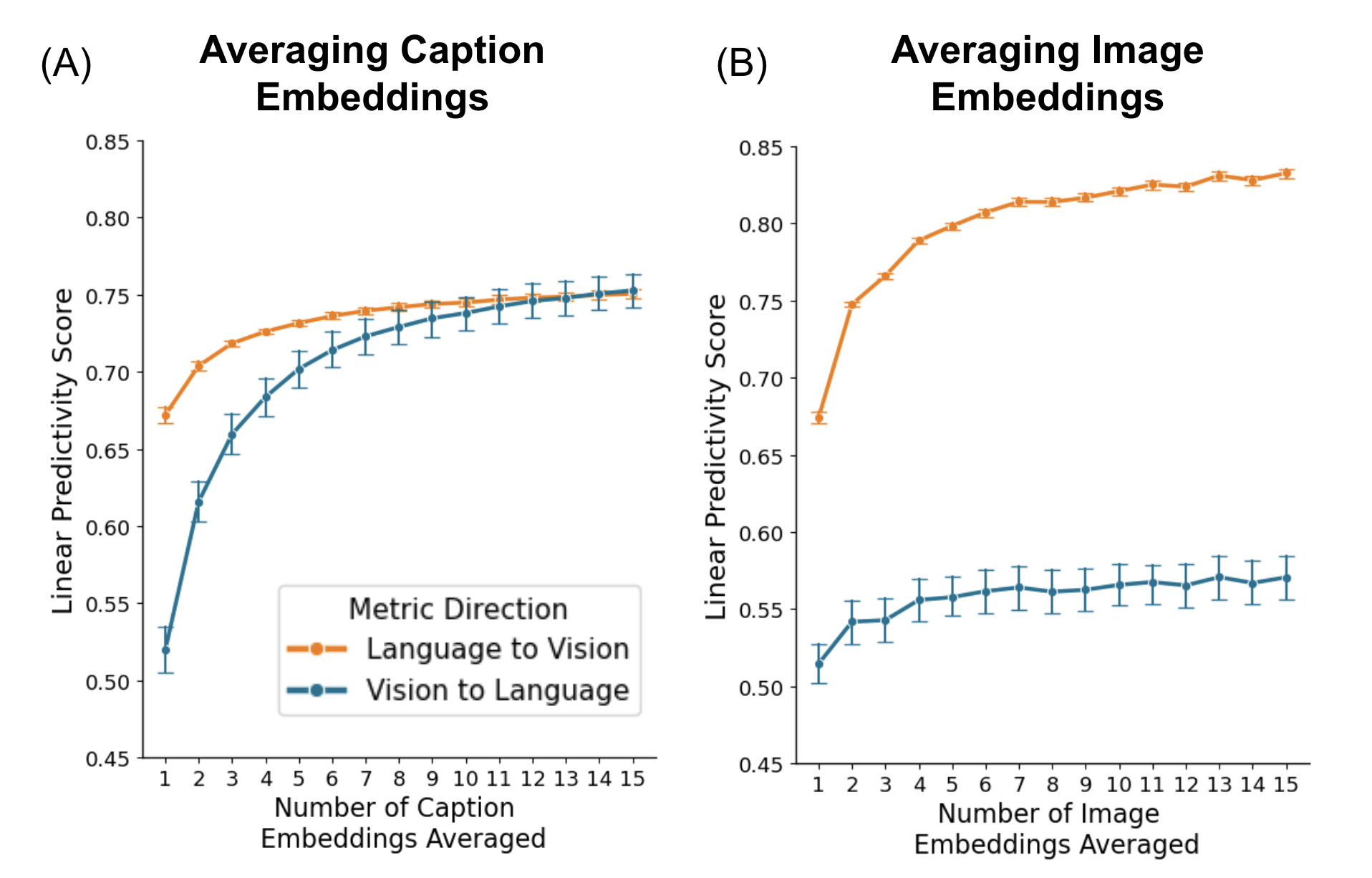}
\caption{Effect of aggregation on alignment. Cross-modal aggregation: Averaging (\textbf{A}) multiple caption embeddings for the same image or (\textbf{B}) multiple image embeddings for the same caption steadily increases language–vision and vision–language predictivity. Error bars denote standard error across all model pairs.  
}
\label{fig:avg}
\end{center}
\end{figure}

To quantify the impact of aggregating caption representations, we progressively averaged embeddings from an increasing number of MS-COCO captions per image and computed cross-modal alignment scores. As shown in Figure ~\ref{fig:avg}A, alignment improved monotonically with each additional caption. To locate the point of diminishing returns, we expanded the caption pool by paraphrasing each of the five human-authored captions with Gemini-2.5-Flash (Figure ~\ref{fig:data}C, see Appendix~\ref{app:gemini} Table \ref{tab:prompt} for prompt), creating up to 15 captions per image. In the vision-to-language mapping, alignment continued to rise until roughly ten captions were included, after which the curve plateaued.

We performed the complementary analysis in the opposite direction by synthesizing up to 15 naturalistic images per caption with Stable Diffusion (Figure ~\ref{fig:data}D). Similar to caption aggregation, increasing the number of aggregated image embeddings further improved the alignment (Figure ~\ref{fig:avg}B). The alignment gain is larger when predicting vision from language, and plateaued around seven images.

To confirm that these improvements reflect enhanced semantic information rather than a generic averaging artifact, we repeated both analyses after randomly shuffling the image–caption correspondences (Appendix~\ref{app:baseline}, Figure \ref{fig:shuffled_baseline}). Under this mismatch baseline, embedding aggregation showed no benefit, demonstrating that the effect depends on semantically matched pairs.

We also observe a clear directional asymmetry both analyses: averaging captions benefited vision-to-language predictions, whereas averaging images benefited language-to-vision predictions. This pattern suggests that aggregation may suppress modality-specific noise within the averaged domain, exposing a cleaner semantic signal that is more easily mapped by the other modality.


\begin{figure*}[t]
\begin{center}
\includegraphics[width=0.92\textwidth]{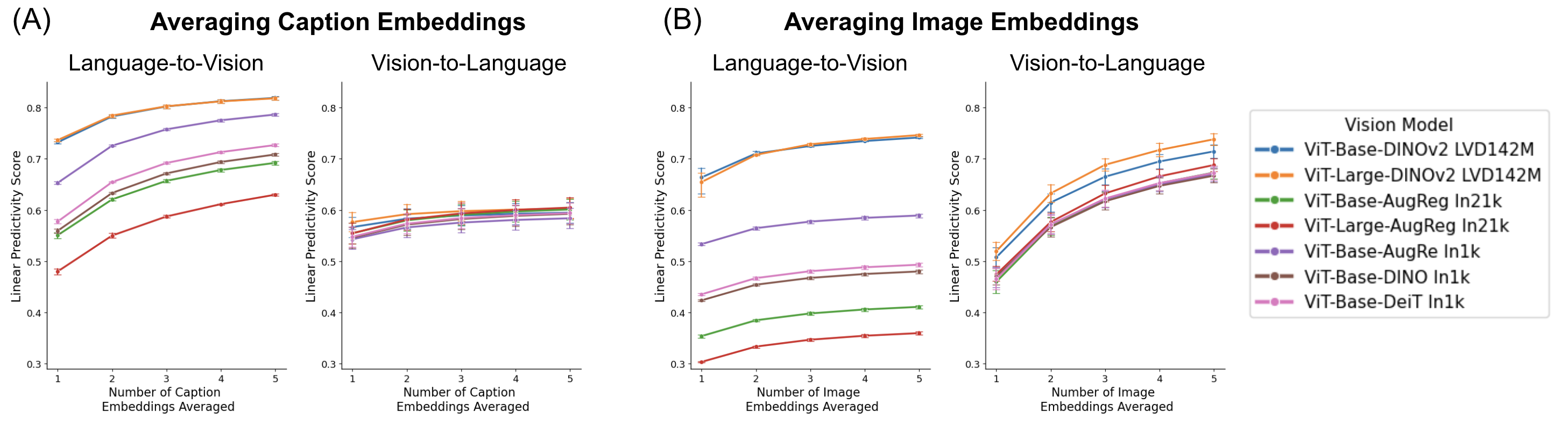}
\caption{Comparing the alignment of different vision models with language models after averaging \textbf{(A)} caption embeddings and \textbf{(B)} image embeddings. }
\label{fig:vision_models}
\end{center}
\end{figure*}

\subsection{Effect of vision models on vision-language alignment}

To assess the generalizability of our findings, we repeated the analyses on seven ViT backbones that differ in objective (strong AugReg, DINO, large-scale DINOv2, supervised distillation DeiT), data scale (ImageNet-1k vs. ImageNet-21k vs. LVD-142 M), and model size (ViT-B/14, ViT-B/16, ViT-L/14, ViT-L/16).

We observe that the improvement of averaging caption embeddings is generalized across different vision model backbones (Figure ~\ref{fig:vision_models}). Notably, when mapping language features into visual space, the alignment differences scores across ViTs were noticeably larger than in the reverse direction. Furthermore, both training methods and data size appear to affect the alignment. When the model size and data were held constant (ViT-B/16, ImageNet-1k), AugReg produced higher alignment than either DINO or DeiT. Keeping the objective similar but increasing the dataset (DINO-ImageNet1k to DINOv2-LVD142m) improved alignment further. However, a larger dataset did not help the AugReg model: its ImageNet-21k checkpoint aligned worse than its ImageNet-1k counterpart. Our current experiment cannot cleanly disentangle the interaction between objective and data distribution. A systematic experiment would be needed to clarify such interactive effects.



\section{Discussion}
Our results provide new evidence that purely uni-modal vision and language models gravitate toward a common semantic manifold. Alignment (i) peaks in their mid-to-late layers where abstract semantic processing occurs, (ii) reduces when we remove or scramble semantic content but survives appearance-only changes, and (iii) exhibits striking correspondence in fine-grained evaluation scenarios with human judgements (e.g., when comparing alignment scores for multiple candidate images corresponding to the same linguistic expression, the model aligns most strongly with the image humans rate as most semantically congruent with the text, and reciprocally for multiple linguistic descriptions of the same image), and (iv) is markedly enhanced when averaging representations corresponding to the same concept in each modality. 
Together, these findings refine the emerging ``Platonic'' view of cross-modal representation: the two modalities do not merely share coarse alignment but capture fine semantic gradients that track human judgments. Our work bridges cognitive science and machine learning by suggesting that a shared code for meaning can emerge implicitly in unimodal systems, even without cross-modal training.

Our work opens several promising avenues for future research. Future studies should investigate how alignment strength varies across different types of visual and linguistic content. Are concrete concepts (e.g., ``dog'', ``chair'') more strongly aligned than abstract concepts (e.g., ``freedom'', ``justice'')? Understanding these variations could reveal fundamental constraints on cross-modal convergence. Different image types—photographs, illustrations, diagrams, artistic renderings—may exhibit varying degrees of alignment with language. Examining these differences could illuminate how visual style and abstraction influence semantic encoding and cross-modal correspondence. 

Our discovery that alignment strengthens when averaging concept-specific representations raises intriguing questions about the geometric properties of these embeddings. Future work should explore whether averaging acts as a denoising mechanism that preserves core semantic content while reducing modality-specific variations. Additionally, it would be interesting to investigate whether averaging techniques applied to paraphrases of the same linguistic expression could enhance performance on downstream tasks involving natural language inference. Prompt ensembling in CLIP—where averaging multiple engineered text prompts boosts zero-shot accuracy in a multimodal model~\citep{radford2021learning}—offers a useful parallel.

While our study demonstrates alignment at the representation level, identifying which specific features or dimensions drive this alignment remains an open question. Future research should develop techniques to isolate the most aligned dimensions between vision and language models and analyze their semantic properties.

Further, investigating how alignment patterns evolve during training could provide insights into the developmental trajectory of cross-modal correspondence. Do alignment patterns appear early in training and strengthen over time, or do they emerge suddenly after sufficient exposure to domain-specific data? This temporal perspective could reveal fundamental insights about how semantic convergence develops in neural networks trained on different modalities.

\section*{Limitations}
Our analysis primarily relies on linear predictivity, complemented by CKA to verify robustness. While these provide complementary perspectives, they still represent only a subset of possible approaches to assessing representational alignment. 
Future work could benefit from employing a broader spectrum of alignment metrics to provide a more complete understanding of vision-language relationships. For instance, more constrained mapping approaches—such as orthogonal transformations in Procrustes analysis~\citep{williams2021generalized} or permutation-based methods like permutation score and soft matching score~\citep{khosla2024soft}—might reveal unit-level correspondences between visual and language model representations that linear regression or CKA cannot capture. Other families of metrics, including Representational Similarity Analysis~\citep{kriegeskorte2008representational} for population-level relationships and neighborhood-based approaches (e.g., mutual k-NN) for local structure, could further enrich the picture.These complementary metrics would provide a multi-faceted view of the nature of alignment between vision and language models. Our analysis does not fully reveal which specific features drive the observed alignment between vision-only and language-only models, nor does it identify the scenarios where these models systematically diverge in their representations. Investigating these questions would require more extensive probing of representations across diverse stimuli and large-scale datasets.



The synthetic nature of our image dataset introduces another limitation. While diffusion models generate high-quality images corresponding to text prompts, some generated images may not perfectly capture the semantic content or nuances present in the texts. This potential mismatch between text and generated images could influence our alignment measurements and subsequent interpretations.

Furthermore, our work examines models trained at a specific point in time, with particular architectures and training objectives. As model architectures and training paradigms evolve, the nature of cross-modal alignment may change significantly.

Finally, representational similarity is descriptive. It does not prove shared processing mechanisms or functional interchangeability. Causal interventions are needed to determine whether the aligned dimensions are necessary for each model's downstream behavior.

\bibliography{custom}

\newpage 

\section{Appendix}

\appendix

\section{Methods}

\subsection{Alignment metrics: implementation details.} 
\label{app:metrics}
\paragraph{Linear Predictivity.}  
For each pair of vision model and language model, we construct $\mathbf{X}\!\in\!\mathbb{R}^{N\times d_X}$ and $\mathbf{Y}\!\in\!\mathbb{R}^{N\times d_Y}$ where $N=1,000$ image–caption combinations in a dataset (e.g. MS-COCO-val2017 or Pick-a-Pic), $d_X$ is the language-feature dimensionality (e.g.\ 2560 for BLOOMZ-3B), and $d_Y$ is the vision-feature dimensionality (e.g.\ 1024 for ViT-Large14-DINOv2). 

A ridge map is fit with 5-fold cross-validation (outer KFold with shuffling) to pick the best $\lambda$ on the training data, where 
$\lambda \in \{10^{-8},10^{-7},\ldots,10^{8}\}$ (17 logarithmically spaced
values). 
Features are z-scored using training statistics and the transform is applied to the test split. 

On the test split, $\hat{\mathbf{Y}}=\mathbf{X}\mathbf{\hat W}$ is evaluated by Pearson $r$ between $\hat{\mathbf{Y}}$ and ${\mathbf{Y}}$, averaged over target dimensions; the final score is the mean across folds. Both directions are reported ( $\mathbf{X}\!\to\!\mathbf{Y}$ and
$\mathbf{Y}\!\to\!\mathbf{X}$).

\paragraph{Centered Kernel Alignment (CKA).}

CKA \cite{kornblith2019cka} measures similarity between representational spaces in a way that is invariant to isotropic scaling and orthogonal transformations, and is symmetric between modalities. We applied CKA to the same vision–language layer pairs as in the linear predictivity analysis. 

For each vision–language model pair, item$\times$feature matrices 
$\mathbf{X}\!\in\!\mathbb{R}^{N\times d_X}$ and 
$\mathbf{Y}\!\in\!\mathbb{R}^{N\times d_Y}$ are constructed over the same held-out items.
With kernels $\kappa_X,\kappa_Y$ and Gram matrices $K_{ij}=\kappa_X(x_i,x_j)$ and 
$L_{ij}=\kappa_Y(y_i,y_j)$, CKA is computed via the (biased) HSIC normalization:
\begin{align}
\mathrm{CKA}(K,L)
&= \frac{\mathrm{HSIC}(K,L)}
        {\sqrt{\mathrm{HSIC}(K,K)\,\mathrm{HSIC}(L,L)}},\\
\mathrm{HSIC}(K,L)
&= \tfrac{1}{N^{2}}\operatorname{tr}(\tilde{K}\,\tilde{L}),
\end{align}
where $\tilde{K}=HKH$, $\tilde{L}=HLH$, and 
$H=I-\tfrac{1}{N}\mathbf{1}\mathbf{1}^{\!\top}$ is the centering matrix.
In the \emph{linear} case used here, $\kappa_X(u,v)=u^\top v$ and $\kappa_Y(u,v)=u^\top v$,
so $K=\mathbf{X}\mathbf{X}^{\!\top}$ and $L=\mathbf{Y}\mathbf{Y}^{\!\top}$ (double-centered by $H$ inside HSIC).

\bigskip

\subsection{Caption manipulation procedure}

Captions are tokenized and \emph{part-of-speech (POS)} tagged using spaCy (\texttt{en\_core\_web\_sm}). We use spaCy’s Universal POS (UPOS) labels (\texttt{Token.pos\_}) to create two filtered variants per caption: \textbf{N} keeps only tokens labeled \{\texttt{NOUN}\}, and \textbf{NV} keeps \{\texttt{NOUN}, \texttt{VERB}\}. All other tokens are removed, and the remaining tokens are rejoined with single spaces.

A scrambled baseline (random permutation of the word tokens within a caption, fixed seed) is implemented separately.




\label{app:caption_manipulation}

\subsection{MS-COCO caption generation.} 
\label{app:gemini}

Caption paraphrases for MS-COCO were generated using Gemini-2.5-Flash to support the embedding-averaging analyses (§3.4). For each image, the five human captions are provided as context, and the model is asked to produce 10 new captions that preserve meaning while varying wording and surface form (Table~\ref{tab:prompt}).

\begin{table}[htbp]
  \centering
  \begin{tcolorbox}[width=0.48\textwidth, boxrule=0.5pt, arc=0pt, auto outer arc]
    \centering
    \textbf{Gemini-2.5-Flash Prompt}\\[1ex]
    \begin{lstlisting}[basicstyle=\ttfamily\tiny, breaklines=true, columns=flexible, frame=none]
Prompt:

f"""You are an expert image captioner. I'll show you some existing captions for an image, and your task is to generate 10 NEW captions that:
1. Are similar in style and detail level to the existing captions
2. Capture the same meaning but with different wording
3. Are direct, concise descriptions (around 10-15 words each)
4. Are worded differently from each existing caption and from each other

Here are the existing captions:
{insert all captions text for the image here}

Generate 10 new captions formatted exactly as:
1. [First new caption]
2. [Second new caption]
3. [Third new caption]
4. [Fourth new caption]
5. [Fifth new caption]
6. [Sixth new caption]
7. [Seventh new caption]
8. [Eighth new caption]
9. [Ninth new caption]
10. [Tenth new caption]"""
    \end{lstlisting}
  \end{tcolorbox}
  \caption{Prompt used to generate MS-COCO caption paraphrases with Gemini-2.5-Flash.}
  \label{tab:prompt}
\end{table}

\subsection{MS-COCO image generation.}

Synthetic images for MS-COCO are generated with the Diffusers
\texttt{StableDiffusionPipeline} initialized from \texttt{CompVis/stable-diffusion-v1-4}. Each caption text of an MS-COCO image is used as the prompt and \(K{=}2\) variants are sampled per caption with \(\texttt{num\_inference\_steps}=50\), yielding 10 synthesized images per MS-COCO image.

\begin{figure*}[t]
\begin{center}
\includegraphics[width=0.96\textwidth]{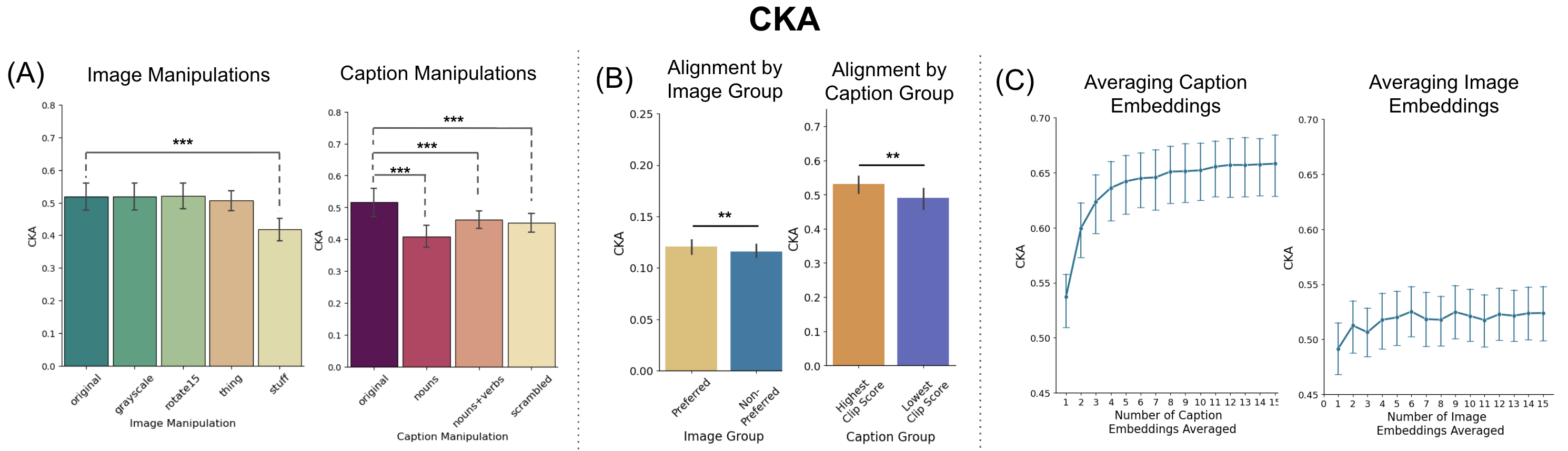}
\caption{ CKA replication across analyses.
\textbf{(A)} Manipulations on images (left) and captions (right). 
\textbf{(B)} Preference. Left: Pick-a-Pic pairs, preferred vs.\ non-preferred. Right: MS-COCO comparison of the highest- vs.\ lowest-CLIP-similarity caption group.
\textbf{(C)} Embedding averaging. }
\label{fig:cka}
\end{center}
\end{figure*}

\section{Extended analyses on MS-COCO with Linear Predictivity}

\subsection{Baseline alignment with shuffled image-caption correspondences.} 
\label{app:baseline}

\begin{figure}[ht]
\begin{center}
\includegraphics[width=0.48\textwidth]{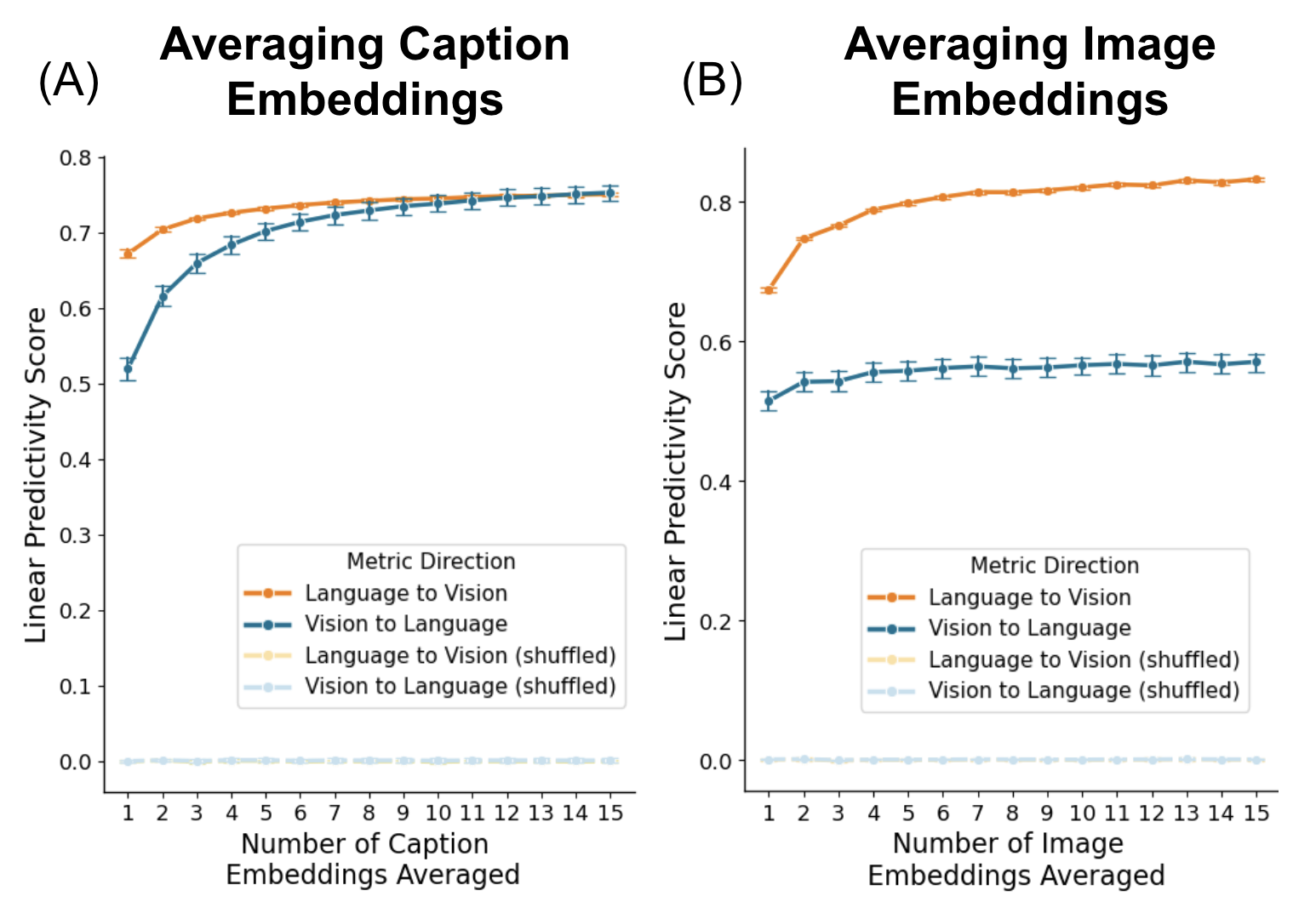}
\caption{Effect of aggregation on alignment with a mismtach baseline.
}
\label{fig:shuffled_baseline}
\end{center}
\end{figure}

Under the image-caption mismatch baseline, averaging multiple embeddings does not improve vision-language alignment: the alignment score remains around 0 in both mapping directions (Figure \ref{fig:shuffled_baseline}).

\subsection{Embedding aggregation effect on manipulated captions.} 
\label{sec:appendix}

\begin{figure}[ht]
\begin{center}
\includegraphics[width=0.48\textwidth]{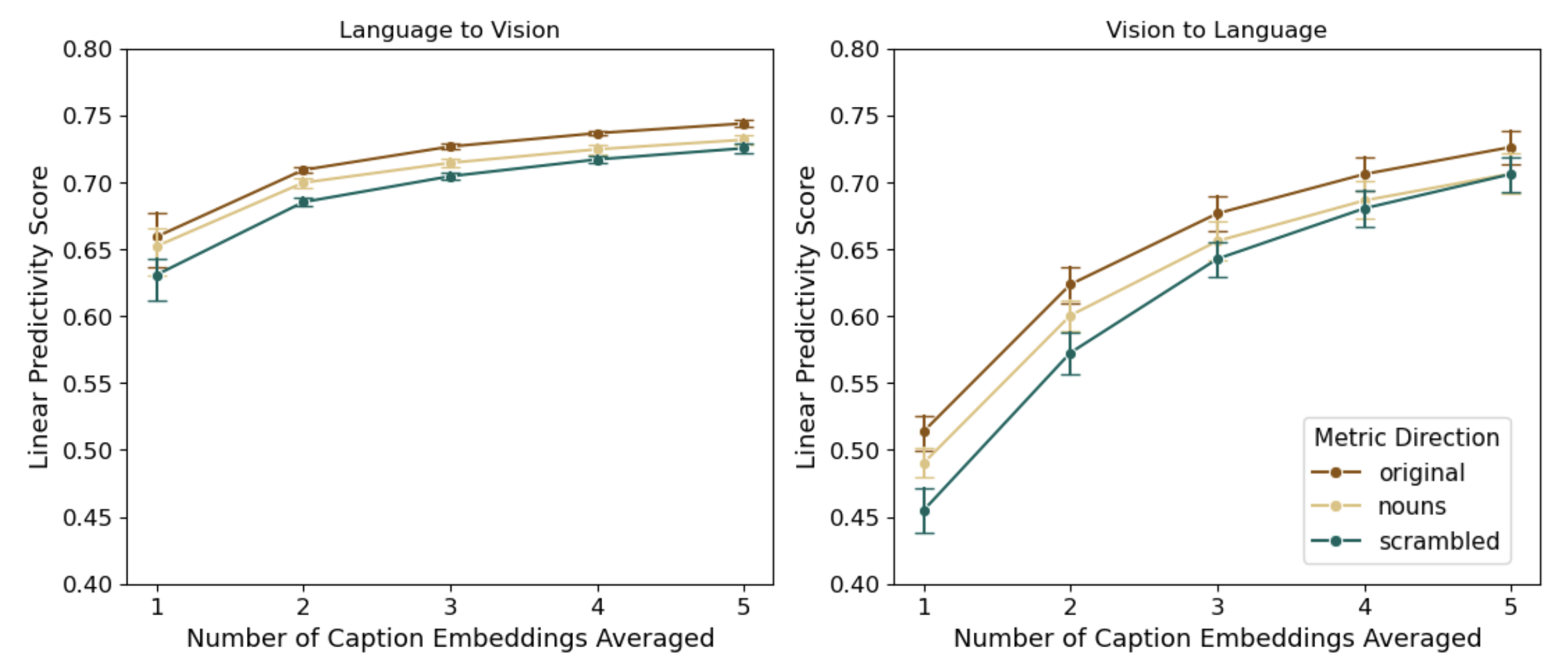}
\caption{Effect of aggregation on alignment with manipulated captions which either only includes nouns or are scrambled in word order.
}
\label{fig:avg_manipulated_captions}
\end{center}
\end{figure}

Given that averaging caption embeddings enhances vision-language alignment, we also explored whether the embeddings of semantically manipulated captions would also benefit from embedding aggregation (Figure \ref{fig:avg_manipulated_captions}). Interestingly, the alignment was enhanced even though the embeddings come from manipulated captions.

\section{Alternative metric: CKA}

\label{app:cka}

To assess metric robustness, we replicated our core analyses with linear CKA on the same held-out items, models, and layers. Because CKA is symmetric, it does not encode the L$\rightarrow$V vs.\ V$\rightarrow$L directionality.

CKA reproduces the qualitative patterns (Figure \ref{fig:cka}A-C). For captions, semantic disruptions (nouns-only, nouns+verbs, and scrambled) reduce alignment, and for images, retaining only stuff (things masked out) yields a significant decrease relative to originals, whereas retaining only things (stuff masked out) shows a smaller, non-significant decrease (Figure \ref{fig:cka}A). Human-preferred pairs have higher alignment (Figure \ref{fig:cka}B). Lastly, averaging embeddings increases alignment, although the gain is weaker yet persistent for image-embedding averaging (Figure \ref{fig:cka}C). 


\section{Additional datasets: Flickr8k}
\label{}

\label{app:addl-datasets-brief}

Using the same procedure as the main text (ridge-based linear predictivity), we test dataset generalization on an additional dataset—\textbf{Flickr8k} \citep{hodosh2013flickr8k}—a captioning dataset of 8{,}000 images, each annotated with five human-authored captions \citep{kaggleFlickr8k, hodosh2013flickr8k}. Because Flickr8k lacks instance-level segmentations, we replicate only the core analyses that do not require foreground/background masks.

We randomly sample 1{,}000 images (and their associated five captions) for analysis. We also generate caption paraphrases and synthesized image variants using the same protocols as in the main text to evaluate embedding averaging.

\begin{figure}[ht]
\begin{center}
\includegraphics[width=0.4\textwidth]{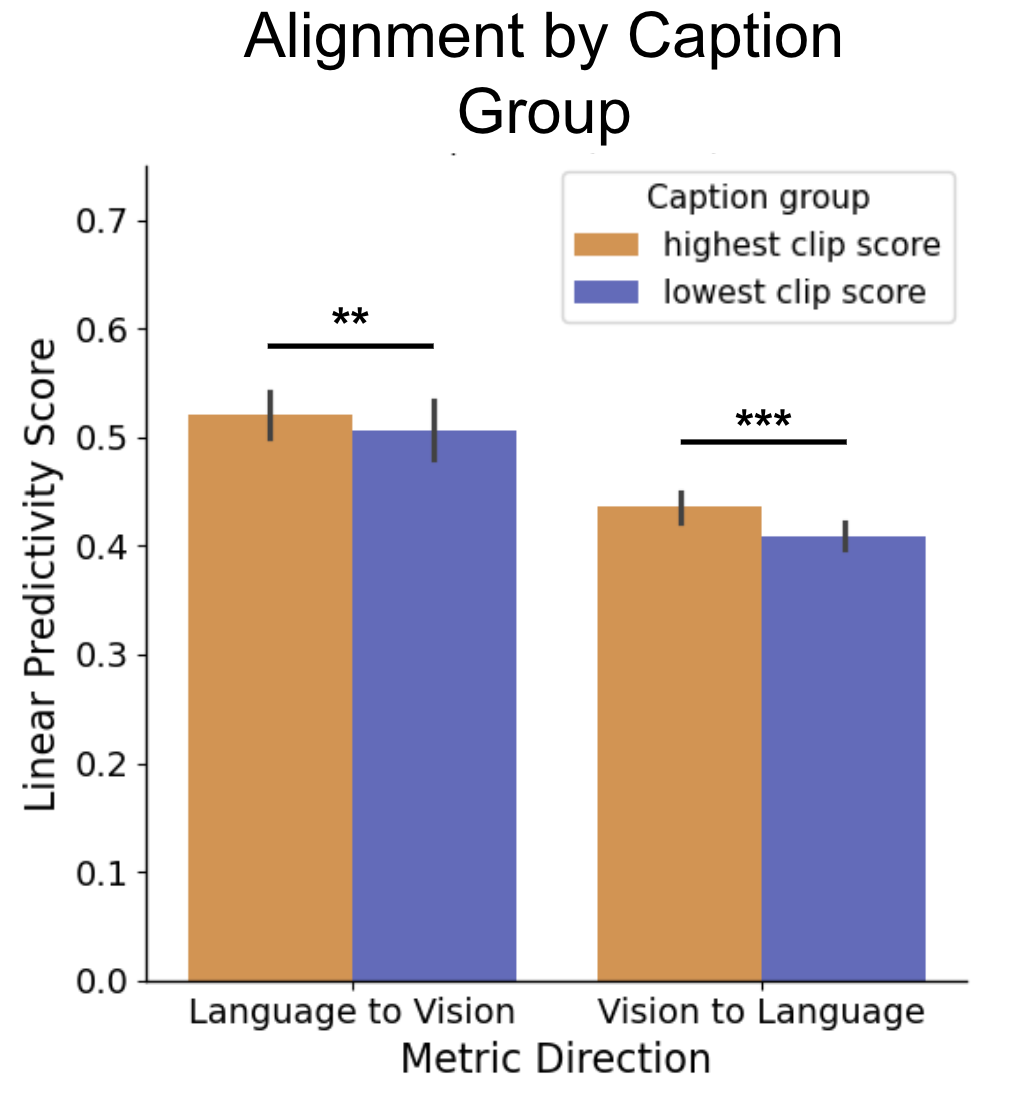}
\caption{Flickr8k dataset linear predictivity scores grouped by caption variation based on CLIP Scores.
}
\label{fig:flickr8k_preference}
\end{center}
\end{figure}

\begin{figure}[ht]
\begin{center}
\includegraphics[width=0.48\textwidth]{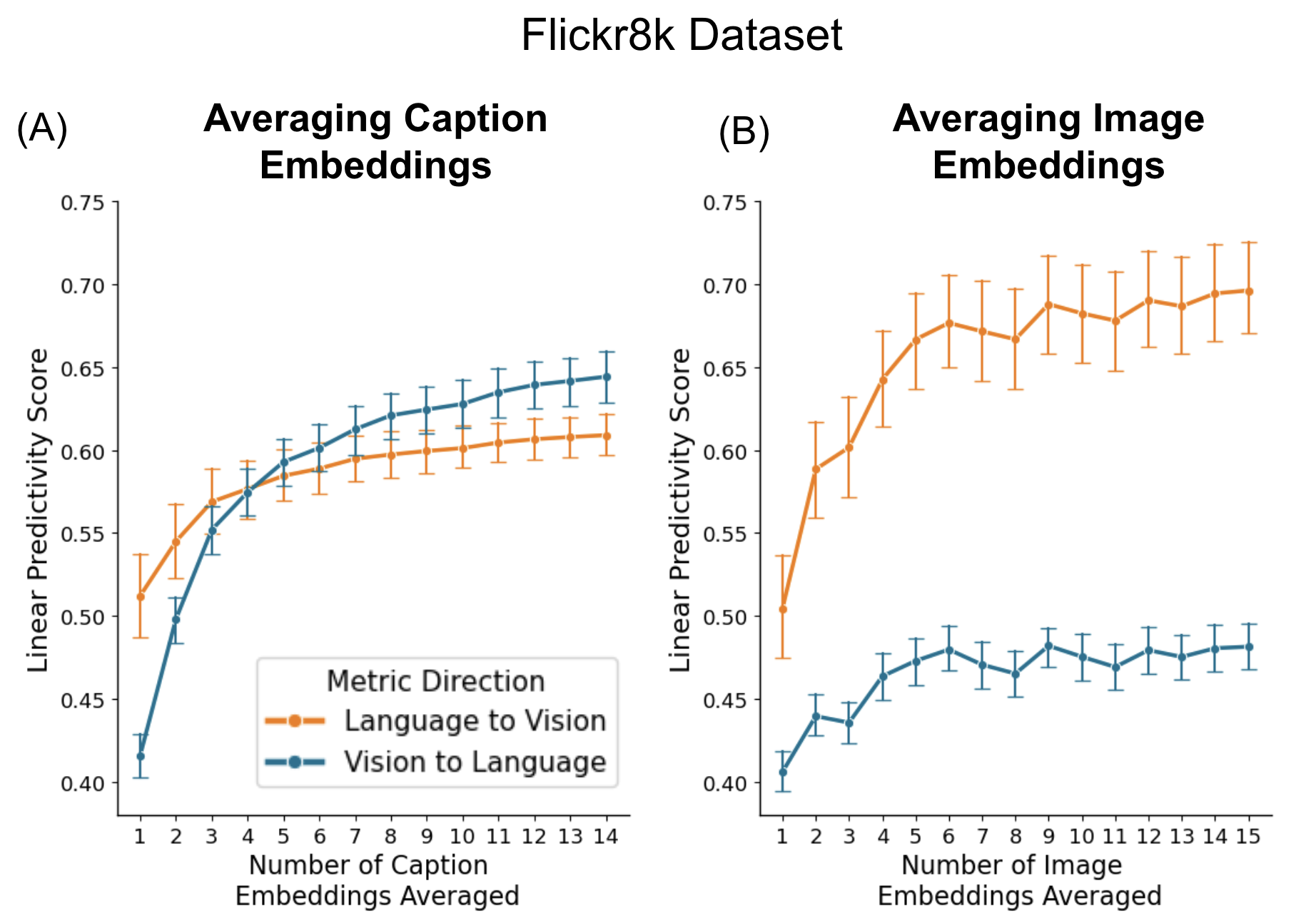}
\caption{Effect of aggregation on alignment on Flickr8k.
}
\label{fig:flickr8k_avg}
\end{center}
\end{figure}

\begin{figure*}[t]
  \centering
  \includegraphics[width=0.85\textwidth]{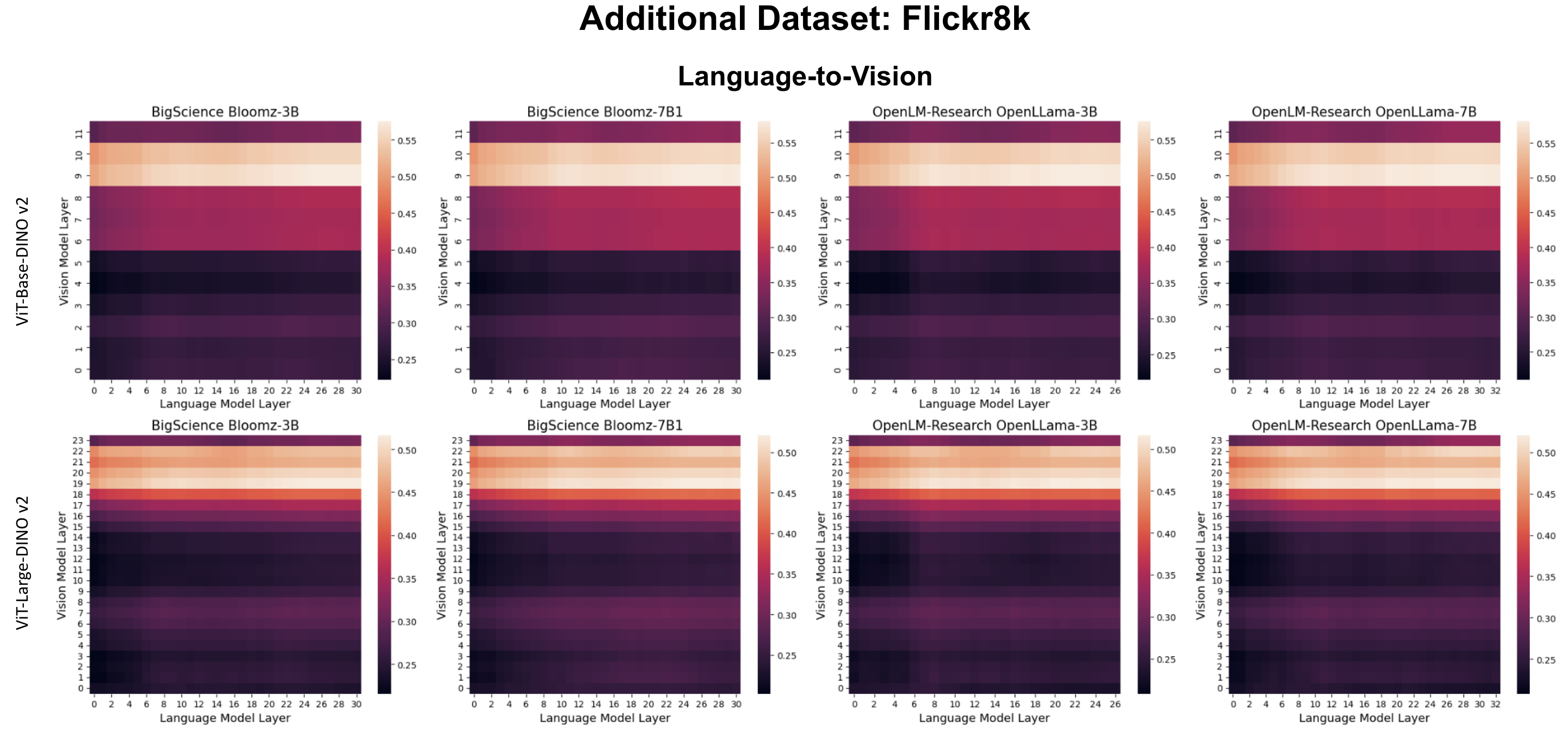}\par\vspace{0.6em}
  \includegraphics[width=0.85\textwidth]{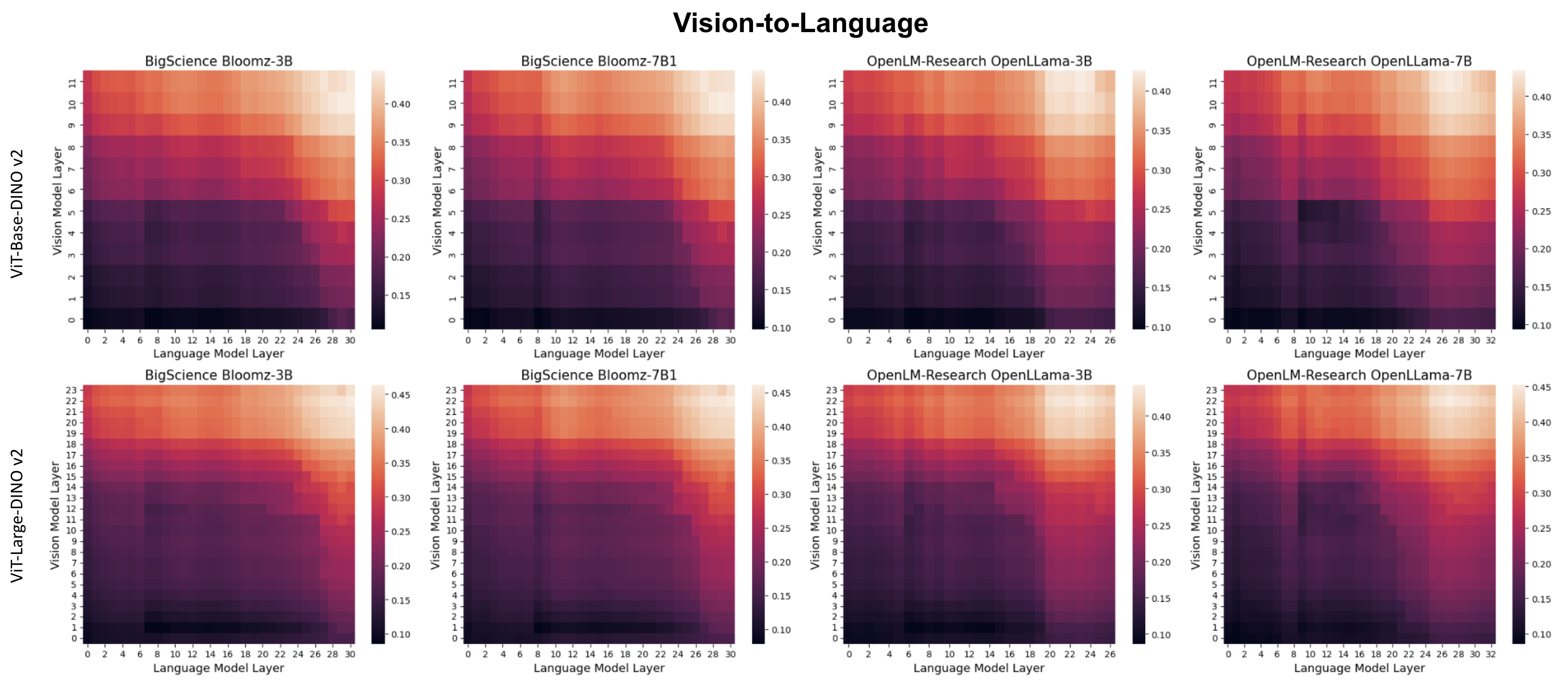}
  \caption{Layer-wise alignment on additional dataset Flickr8k. Top two rows: Alignment computed in language-to-vision direction. Bottom two rows: Alignment computed in vision-to-language direction.}
  \label{fig:heatmap_flickr8k}
\end{figure*}

The main patterns replicate for Flickr8k: alignment increases from mid to late layers and the L$\!\rightarrow$V $>$ V$\!\rightarrow$L asymmetry holds (Figure \ref{fig:heatmap_flickr8k}). The CLIP-based preference proxy yields higher alignment for higher-ranked captions (paired t-test, L$\!\rightarrow$V: $t(7)=5.0520$, $p=0.0015$; V$\!\rightarrow$L: $t(7)=13.8867$, $p<0.0001$; Figure \ref{fig:flickr8k_preference}); and averaging multiple caption/image embeddings improves alignment and plateaus as the number of embeddings increases (Figure \ref{fig:flickr8k_avg}). 


\begin{figure*}[t]
\begin{center}
\includegraphics[width=0.96\textwidth]{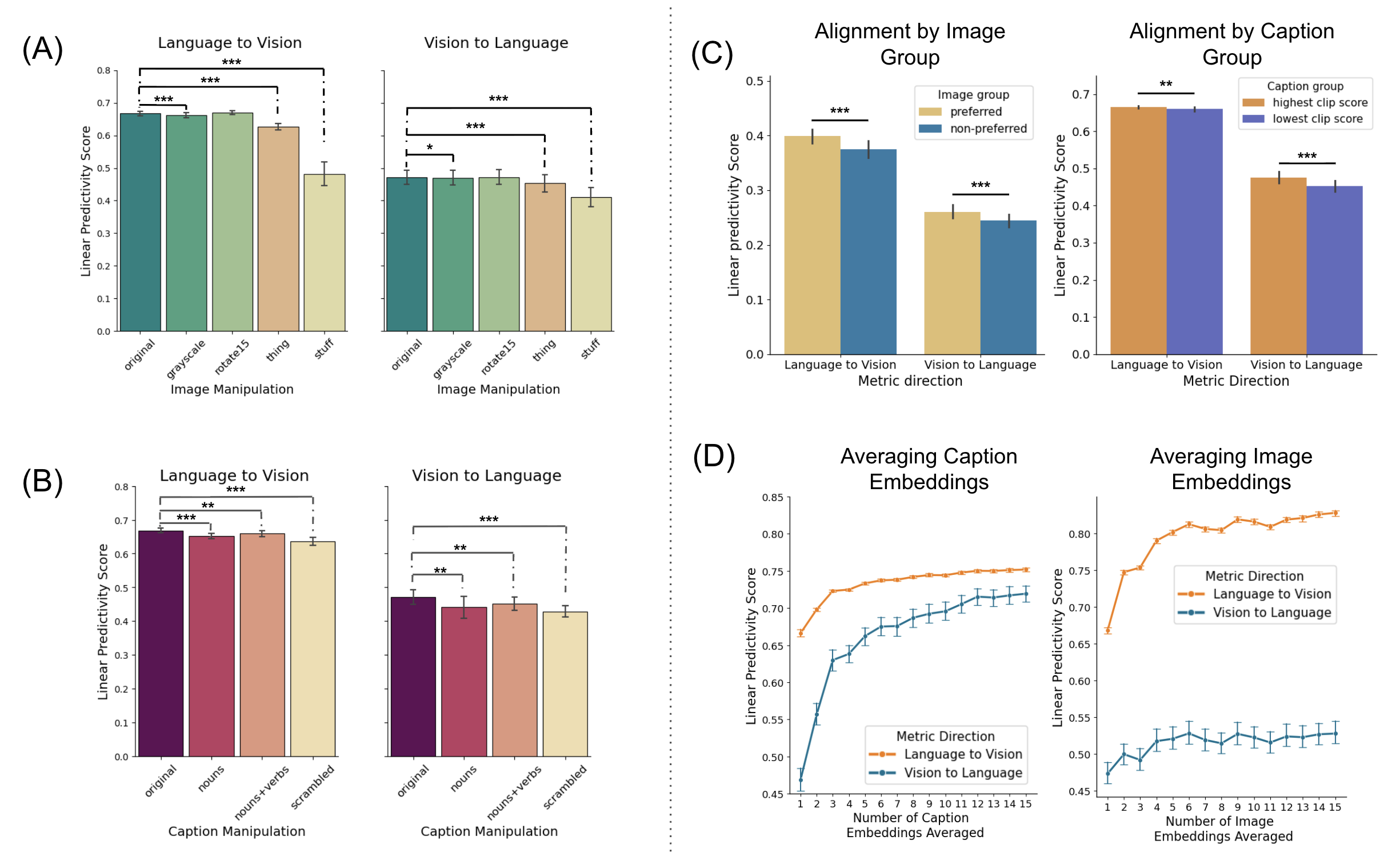}
\caption{ Analysis replication on additional LLM families. 
\textbf{(A)} Image Manipulations.  \textbf{(B)} Caption Manipulations. 
\textbf{(C)} Preference. Left: Pick-a-Pic pairs, preferred vs.\ non-preferred. Right: MS-COCO comparison of the highest- vs.\ lowest-CLIP-similarity caption group.
\textbf{(D)} Embedding averaging. }
\label{fig:llmfam-overview}
\end{center}
\end{figure*}

\section{Expanded LLM families: Qwen, Phi-3, SmolLM}

\label{app:expanded-llm}

Using the same procedure as the main text, we evaluate three additional LLM families—Qwen (3B, 7B) \cite{yang2024qwen2_5}; Phi-3 (mini) \cite{abdin2024phi3}, and SmolLM (1.7B) \cite{allal2025smollm2}—on the same analyses (Figure~\ref{fig:llmfam-overview}A–D). The main patterns replicate across LLM families.

\begin{figure*}[t]
  \centering
  \includegraphics[width=0.85\textwidth]{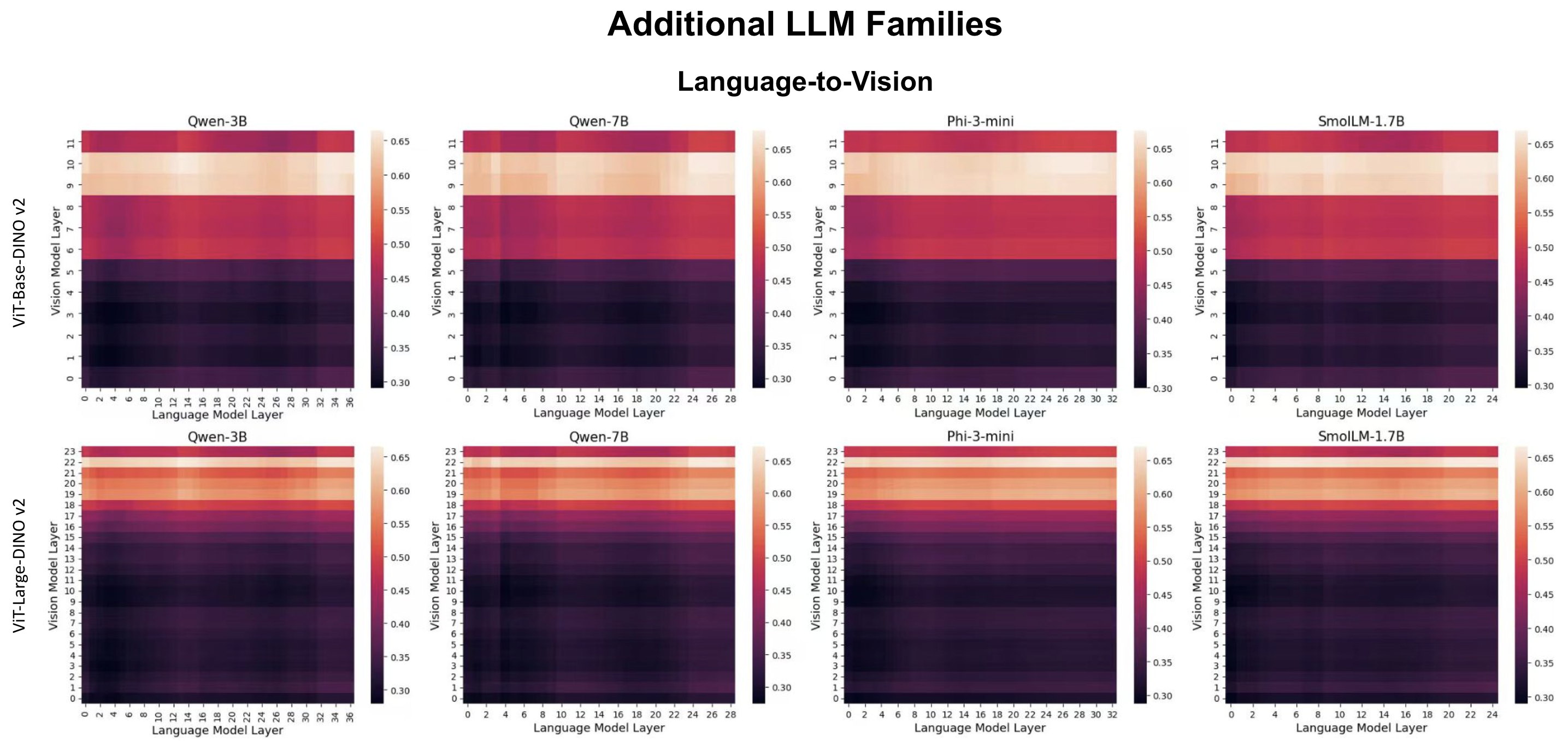}\par\vspace{0.6em}
  \includegraphics[width=0.85\textwidth]{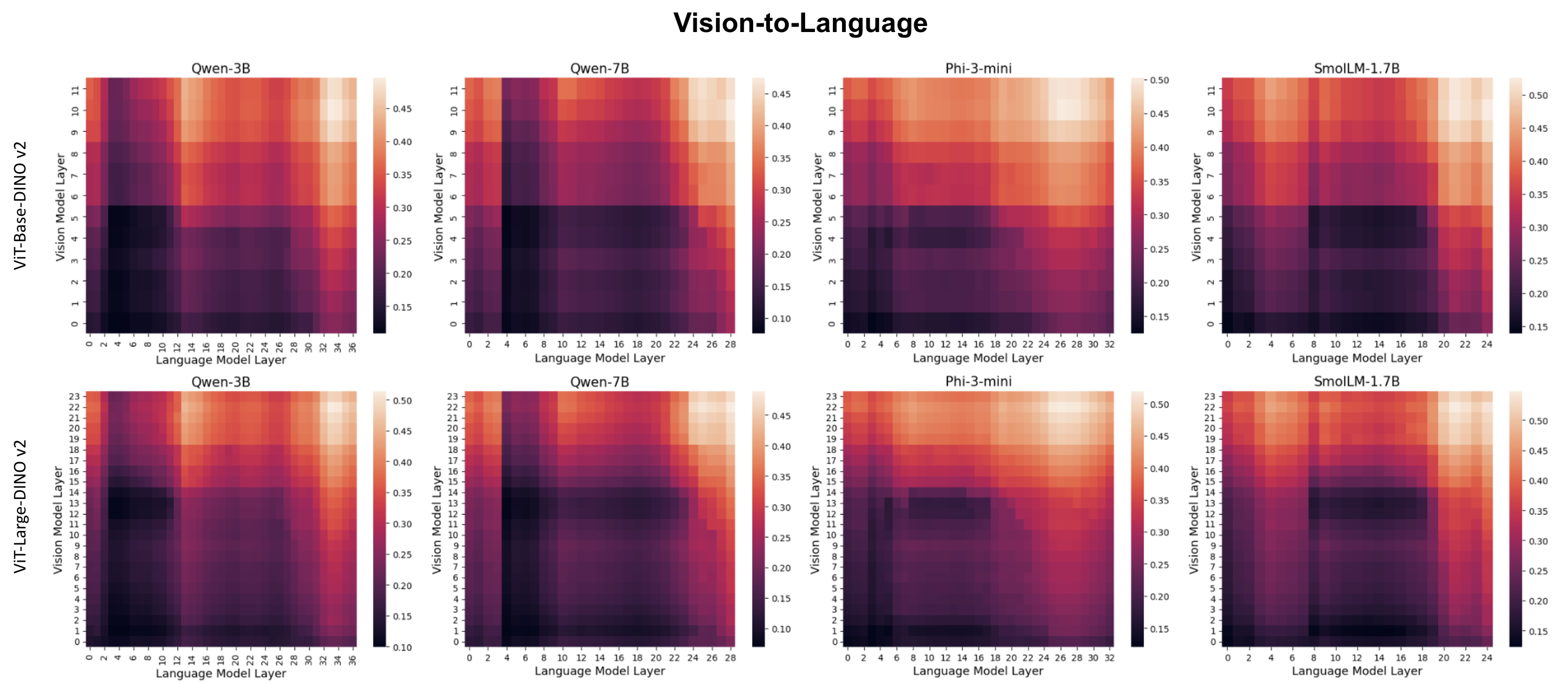}
  \caption{Layer-wise alignment on MS-COCO with additional LLMs: Qwen2.5-3B, Qwen2.5-7B, Phi-3-mini-128k-instruct, and SmolLM2-1.7B. Top two rows: Alignment computed in language-to-vision direction. Bottom two rows: Alignment computed in vision-to-language direction.}
  \label{fig:heatmap_new_llm}
\end{figure*}

\textbf{Layer-wise alignment.} The mid-to-late layer rise holds across families, and the L$\!\rightarrow$V $>$ V$\!\rightarrow$L asymmetry is preserved (Figure ~\ref{fig:heatmap_new_llm}).

\textbf{Image manipulations.} Retaining only "things" foreground and retaining only "stuff" background both significantly reduce alignment. A small rotation does not yield a reliable decrease, whereas converting images to grayscale produces a small but significant reduction (Figure ~\ref{fig:llmfam-overview}A, grayscale: L$\!\rightarrow$V $t(7)=8.1393$, $p=0.0001$, $q=0.0002$; V$\!\rightarrow$L $t(7)=2.7209$, $p=0.0297$, $q=0.0396$; 
rotate15: L$\!\rightarrow$V $t(7)=-1.1300$, $p=0.2957$, $q=0.3379$; V$\!\rightarrow$L $t(7)=-0.5698$, $p=0.5866$, $q=0.5866$; 
stuff-only: L$\!\rightarrow$V $t(7)=12.8386$, $p<0.0001$, $q<0.0001$; V$\!\rightarrow$L $t(7)=12.1947$, $p<0.0001$, $q<0.0001$; 
things-only: L$\!\rightarrow$V $t(7)=21.4355$, $p<0.0001$, $q<0.0001$; V$\!\rightarrow$L $t(7)=5.5535$, $p=0.0009$, $q=0.0014$). A plausible interpretation is that color carries captioned semantics (e.g., color words in text), and these families are sensitive to that finer-grained correspondence.

\textbf{Caption manipulations.} All caption manipulations (nouns-only, nouns+verbs, scrambled) produce significantly lower alignment than the original captions in each family (Figure ~\ref{fig:llmfam-overview}B; nouns: L$\!\rightarrow$V $t(7)=6.3039$, $p=0.0004$, $q=0.0008$; V$\!\rightarrow$L $t(7)=5.3781$, $p=0.0010$, $q=0.0015$; 
nouns+verbs: L$\!\rightarrow$V $t(7)=3.6133$, $p=0.0086$, $q=0.0086$; V$\!\rightarrow$L $t(7)=3.6944$, $p=0.0077$, $q=0.0086$; 
scrambled: L$\!\rightarrow$V $t(7)=8.9154$, $p<0.0001$, $q=0.0001$; V$\!\rightarrow$L $t(7)=9.8068$, $p<0.0001$, $q=0.0001$.)
.

\textbf{Human preference / CLIP proxy.} The preference effect replicates across families: preferred $>$ non-preferred image group (paired t-test, L→V: $t(7)=14.0585$, $p<0.0001$; V→L: $t(7)=9.1631$, $p<0.0001$), and high CLIP Score $>$ low CLIP score (paired t-test, L→V: $t(7)=3.5055$, $p=0.0099$; V→L: $t(7)=14.4357$, $p<0.0001$). (Figure ~\ref{fig:llmfam-overview}C)

\textbf{Embedding averaging.} Averaging embeddings increases alignment and plateaus with larger number of examplars for both caption- and image-embedding averaging (Figure ~\ref{fig:llmfam-overview}D).




\begin{figure*}[t]
\begin{center}
\includegraphics[width=0.8\textwidth]{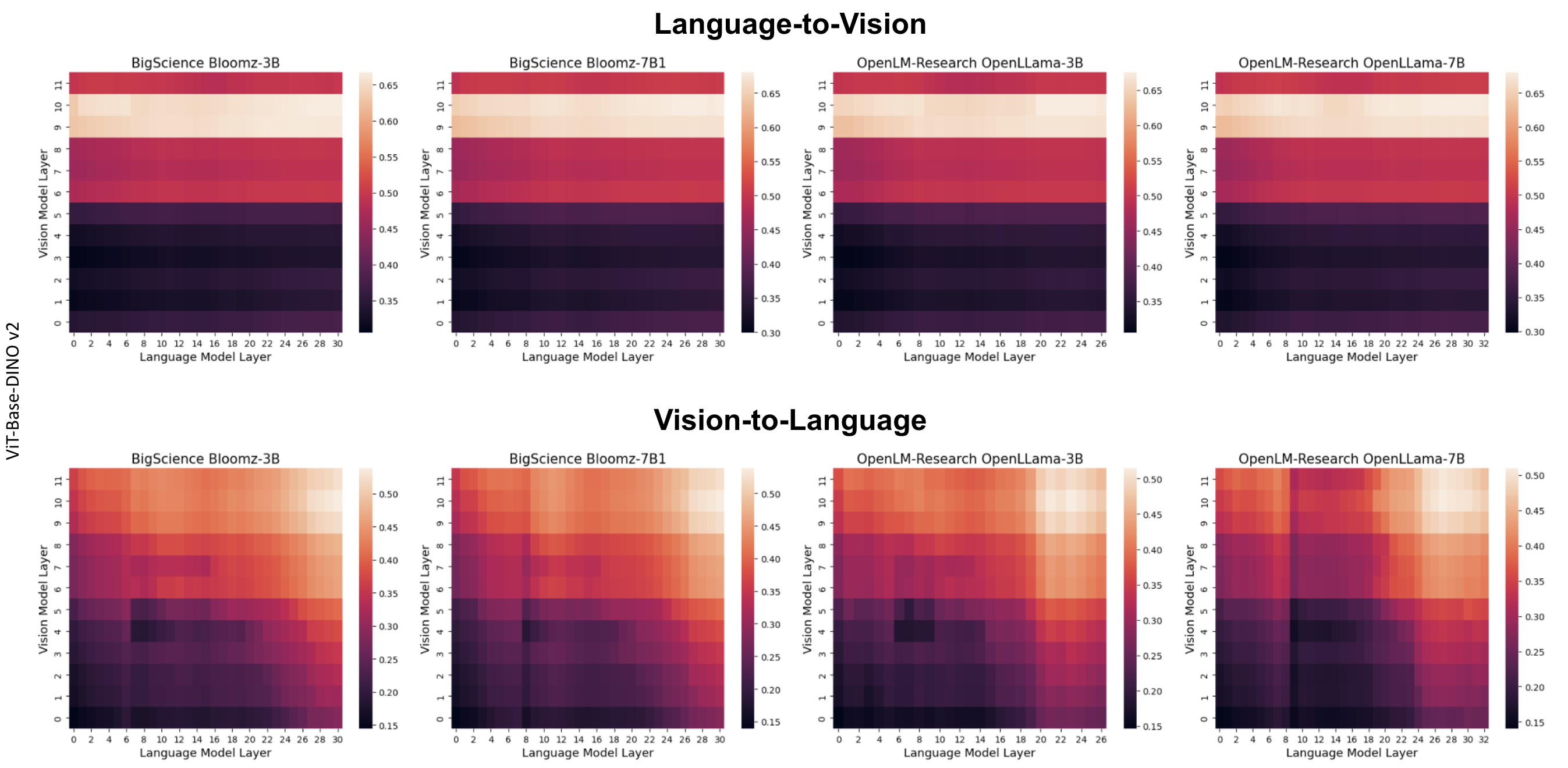}
\caption{Layer-wise alignment for additional vision-language model pairs (with ViT-Base-DINO v2).}
\label{fig:heatmap_new}
\end{center}
\end{figure*}

\section*{Code Availability}
Code and scripts are available at \url{https://github.com/zoewhe/vision-language-alignment}. 

\end{document}